\documentclass[letterpaper, 10 pt, conference]{ieeeconf}  

\IEEEoverridecommandlockouts                              

\usepackage[backend=biber,style=ieee]{biblatex}
\title{\LARGE \bf
\approach: Path Planning via Semantic Lifting in \\ TSDF-Guided Gaussian Splatting
}

\author{Hannah Schieber$^{1,*}$, Dominik Frischmann$^{1,*}$, Victor Schaack$^{1}$, Angela P. Schoellig$^{2}$ and Daniel Roth$^{1}$
\thanks{* equal contribution}
\thanks{$^{1}$Hannah Schieber, Dominik Frischmann, Victor Schaack, and Daniel Roth are with the Technical University of Munich,
Human-Centered Computing and Extended Reality Lab,
TUM University Hospital, Clinic for Orthopedics and Sports Orthopedics,
Munich Institute of Robotics and Machine Intelligence (MIRMI)
        {\tt\small hannah.schieber@tum.de}}%
\thanks{$^{1}$Angela Schoellig is with the Technical University of Munich,
Learning Systems and Robotics Lab,
Munich Institute of Robotics and Machine Intelligence (MIRMI)}%
}

\usepackage{mathptmx}
\usepackage{amsmath}
\usepackage{graphicx}
 \usepackage{colortbl}
 \usepackage{booktabs}
 \usepackage{multirow}
\usepackage[nolist]{acronym}
\usepackage{xspace}
\def\approach{LiftNav\xspace}
\usepackage{subcaption}
\begin{document}
\begin{acronym}[Bspwwww.]  

\acro{ar}[AR]{augmented reality}
\acro{ap}[AP]{average precision}
\acro{api}[API]{application programming interface}
\acroplural{ann}[ANN]{artifical neural networks}
\acro{bev}[BEV]{bird eye view}
\acro{rbob}[BRB]{Bottleneck residual block}
\acroplural{rbob}[BRBs]{Bottleneck residual blocks}
\acro{mbiou}[mBIoU]{mean Boundary Intersection over Union}
\acro{poi}[POI]{Point of Interest}
\acro{cai}[CAI]{computer-assisted intervention}
\acro{ce}[CE]{cross entropy}
\acro{cad}[CAD]{computer-aided design}
\acro{cnn}[CNN]{convolutional neural network}

\acro{crf}[CRF]{conditional random fields}
\acro{dpc}[DPC]{dense prediction cells}
\acro{dla}[DLA]{deep layer aggregation}
\acro{dnn}[DNN]{deep neural network}
\acroplural{dnn}[DNNs]{deep neural networks}
\acro{dbscan}[DBSCAN]{Density-Based Spatial Clustering of Applications with Noise}

\acro{da}[DA]{domain adaption}
\acro{dr}[DR]{domain randomization}
\acro{fat}[FAT]{falling things}
\acro{fcn}[FCN]{fully convolutional network}
\acroplural{fcn}[FCNs]{fully convolutional networks}
\acro{fov}[FoV]{field of view}
\acro{fv}[FV]{front view}
\acro{fp}[FP]{False Positive}
\acro{fpn}[FPN]{feature Pyramid network}
\acro{fn}[FN]{False Negative}
\acro{fmss}[FMSS]{fast motion sickness scale}
\acro{gan}[GAN]{generative adversarial network}
\acroplural{gan}[GANs]{generative adversarial networks}
\acro{gcn}[GCN]{graph convolutional network}
\acroplural{gcn}[GCNs]{graph convolutional networks}
\acro{gs}[GS]{Gaussian Splatting}
\acro{gg}[GG]{Gaussian Grouping}
\acro{gt}[GT]{ground truth}
\acro{hmi}[HMI]{Human-Machine-Interaction}
\acro{hmd}[HMD]{Head Mounted Display}
\acroplural{hmd}[HMDs]{head mounted displays}
\acro{iou}[IoU]{intersection over union}
\acro{irb}[IRB]{inverted residual bock}
\acroplural{irb}[IRBs]{inverted residual blocks}
\acro{ipq}[IPQ]{igroup presence questionnaire}
\acro{iin}[IIN]{Instance ImageGoal Navigation}
\acro{knn}[KNN]{k-nearest-neighbor}
\acro{lidar}[LiDAR]{light detection and ranging}
\acro{lsfe}[LSFE]{large scale feature extractor}
\acro{llm}[LLM]{large language model}
\acro{map}[mAP]{mean average precision}
\acro{mc}[MC]{mismatch correction module}
\acro{miou}[mIoU]{mean intersection over union}
\acro{mis}[MIS]{Minimally Invasive Surgery}
\acro{msdl}[MSDL]{Multi-Scale Dice Loss}
\acro{ml}[ML]{Machine Learning}
\acro{mlp}[MLP]{multilayer perception}
\acro{miou}[mIoU]{mean Intersection over Union}
\acro{nn}[NN]{neural network}
\acroplural{nn}[NNs]{neural networks}
\acro{ndd}[NDDS]{NVIDIA Deep Learning Data Synthesizer}
\acro{nocs}[NOCS]{Normalized Object Coordiante Space}
\acro{nerf}[NeRF]{Neural Radiance Fields}
\acro{NVISII}[NVISII]{NVIDIA Scene Imaging Interface}
\acro{ngp}[NGP]{Neural Graphics Primitives}
\acro{or}[OR]{Operating Room}
\acro{pbr}[PBR]{physically based rendering}
\acro{psnr}[PSNR]{peak signal-to-noise ratio}
\acro{pnp}[PnP]{Perspective-n-Point}
\acro{rv}[RV]{range view}
\acro{roi}[RoI]{region of interest}
\acroplural{roi}[RoIs]{region of interests}
\acro{rbab}[BB]{residual basic block}
\acro{ras}[RAS]{robot-assisted surgery}
\acroplural{rbab}[BBs]{residual basic blocks}
\acro{spp}[SPP]{spatial pyramid pooling}
\acro{sh}[SH]{spherical harmonics}
\acro{sgd}[SGD]{stochastic gradient descent}
\acro{sdf}[SDF]{signed distance field}
\acro{sfm}[SfM]{structure-from-motion}
\acro{sam}[SAM]{Segment-Anything}
\acro{sus}[SUS]{system usability scale}
\acro{ssim}[SSIM]{structural similarity index measure}
\acro{sfm}[SfM]{structure from motion}
\acro{slam}[SLAM]{simultaneous localization and mapping}
\acro{sfc}[SFC]{Safe Flight Corridor}
\acro{tp}[TP]{True Positive}
\acro{tn}[TN]{True Negative}
\acro{thor}[thor]{The House Of inteRactions}
\acro{tsdf}[TSDF]{truncated signed distance function}
\acro{vr}[VR]{Virtual Reality}
\acro{ycb}[YCB]{Yale-CMU-Berkeley}

\acro{ar}[AR]{augmented reality}
\acro{ate}[ATE]{absolute trajectory error}
\acro{bvip}[BVIP]{blind or visually impaired people}
\acro{cnn}[CNN]{convolutional neural network}
\acro{c2f}[c2f]{coarse-to-fine}
\acro{fov}[FoV]{field of view}
\acro{gan}[GAN]{generative adversarial network}
\acro{gcn}[GCN]{graph convolutional Network}
\acro{gnn}[GNN]{Graph Neural Network}
\acro{hmi}[HMI]{Human-Machine-Interaction}
\acro{hmd}[HMD]{head-mounted display}
\acro{mr}[MR]{mixed reality}
\acro{iot}[IoT]{internet of things}
\acro{llff}[LLFF]{Local Light Field Fusion}
\acro{bleff}[BLEFF]{Blender Forward Facing}

\acro{lpips}[LPIPS]{learned perceptual image patch similarity}
\acro{lss}[LSS]{Lift, Splat, Shoot}
\acro{nerf}[NeRF]{neural radiance fields}
\acro{nvs}[NVS]{novel view synthesis}
\acro{mlp}[MLP]{multilayer perceptron}
\acro{mrs}[MRS]{Mixed Region Sampling}

\acro{or}[OR]{Operating Room}
\acro{pbr}[PBR]{physically based rendering}
\acro{psnr}[PSNR]{peak signal-to-noise ratio}
\acro{pnp}[PnP]{Perspective-n-Point}
%
\acro{sus}[SUS]{system usability scale}
\acro{ssim}[SSIM]{similarity index measure}
\acro{sfm}[SfM]{structure from motion}
\acro{slam}[SLAM]{simultaneous localization and mapping}

\acro{tp}[TP]{True Positive}
\acro{tn}[TN]{True Negative}
\acro{thor}[thor]{The House Of inteRactions}
\acro{ueq}[UEQ]{User Experience Questionnaire}
\acro{vr}[VR]{virtual reality}
\acro{who}[WHO]{World Health Organization}
\acro{xr}[XR]{extended reality}
\acro{ycb}[YCB]{Yale-CMU-Berkeley}
\acro{yolo}[YOLO]{you only look once}
\end{acronym} 

\maketitle
\thispagestyle{empty}
\pagestyle{empty}

\begin{abstract}

Autonomous robots in unknown indoor environments require both reliable collision avoidance and object-level understanding. Classical representations such as TSDF support safe planning but lack semantics, while photorealistic methods like Gaussian Splatting (GS) provide rich appearance yet suffer from soft geometry, limiting precise obstacle avoidance.
We present LiftNav, a hybrid navigation framework built on GSFusion’s TSDF+GS dual map, augmented with a real-time pipeline of YOLO-based detection, TSDF-based 3D lifting, and B-spline trajectory optimization. This design enables flexible semantic navigation without dense 3D embeddings. We further introduce a hinge-loss-based collision penalty that improves trajectory smoothness and safety. We evaluate our approach in a simulation using the Replica dataset. Compared against a state-of-the-art radiance field baseline we show a 100\% feasibility rate and shorter trajectories.

\end{abstract}

\section{INTRODUCTION}

Robust navigation in complex 3D environments requires representations that capture geometry, appearance, and semantics, and potentially dynamics \cite{chen2025splat, brunke2025semantically, bogenberger2026did}. Radiance field methods, especially \ac{gs}, enable efficient and photorealistic modeling, but remain weakly integrated with robotic pipelines, often limited to COLMAP-based setups \cite{chen2025splat}. Embedding semantics directly into \ac{gs} improves scene understanding but increases computational cost and training time \cite{schieber_object_2024}. In contrast, combining accurate geometry with lightweight 2D semantic lifting offers a more efficient alternative.

We propose \approach, which combines photorealistic \ac{gs} mapping with precise \ac{tsdf}-based geometry, defers semantics to a lightweight 2D lifting stage, and restricts their use to navigation-relevant regions, enabling efficient and safe navigation under a lightweight policy. Our contributions are:

\begin{itemize}
\item A semantic lifting approach for \ac{tsdf}-based \ac{gs}.
\item A robust and light-weight semantic navigation pipeline built on \ac{tsdf}.
\end{itemize}

\section{RELATED WORK}
\subsection{Hybrid Representations}

ActiveGS \cite{jin2025activegs} couples a \ac{gs} map with a coarse voxel grid to enable spatial reasoning and exploration by targeting under-reconstructed regions. GSFusion \cite{wei2024gsfusion} integrates \ac{tsdf} geometry with \ac{gs} for online RGB-D mapping, using geometric priors to regularize splatting and suppress artifacts. 

\subsection{3D Path Planning within Radiance Fields}

Leveraging high-quality \ac{gs} scene representations for navigation is, for example, done by Adamkiewicz et al. \cite{adamkiewicz2022nerfnav} within \ac{nerf}. Within \ac{gs}, Chen et al. introduced SplatNav \cite{chen2025splat}, a real-time navigation pipeline in \ac{gs} with two modules, \emph{Splat-Plan} and \emph{Splat-Loc}. Splat-Plan constructs safe convex corridors from Gaussian ellipsoids and generates smooth Bézier trajectories at $>$2 Hz, while Splat-Loc performs $\sim$25 Hz pose estimation by matching monocular RGB input to \ac{gs} renderings via a \ac{pnp} formulation. However, they assume a high-quality known map, leaving a gap in scenarios requiring simultaneous exploration, mapping, and navigation. Similarly, GaussNav \cite{lei2024gaussnav} operates on known envrionemnts. It uses a semantic \ac{gs}, assigning 3D instance labels to Gaussians from clustered 2D masks. Target localization relies on exhaustive rendering and feature matching over candidate views, followed by 2D grid-based planning. While effective, this design incurs high computational cost and low update rates. 

\subsection{Semantic Perception and Spatial Projection}

To leverage 2D semantic detections for navigation, the detections must be projected into metric 3D space. \ac{lss} \cite{philion2020lift} lifts pixels into 3D frustums via predicted depth and aggregates them into a unified \ac{bev}. In \ac{gs}, semantics are often embedded into rendering \cite{schieber_object_2024}, increasing training complexity. These methods typically rely on foundation models; while SAM lacks class labels, newer models like SAM3 \cite{carion2025sam} enable language-guided semantics. Lightweight models such as YOLOv8n-seg \cite{yolov8_ultralytics} offer real-time performance. 

\section{METHOD}

\subsection{Semantic Mapping}

\approach extracts and localizes semantic targets in 3D through two primary phases: (i) 2D object detection \cite{yolov8_ultralytics} and (ii) spatio-temporal semantic lifting. 



\begin{figure}[t!]
    \centering
    \includegraphics[width=\columnwidth]{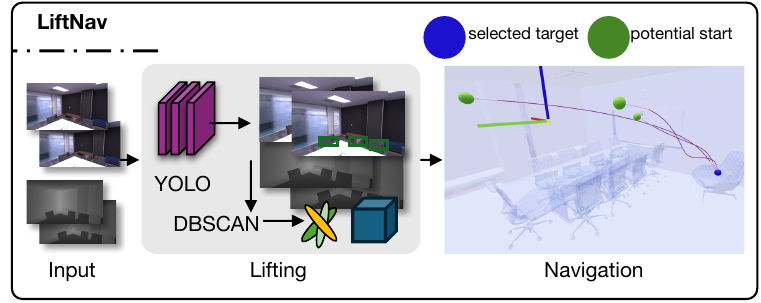}
    \caption{Architecture. We process a RGB-D input stream via GSFusion \cite{wei2024gsfusion}, simultaneously we leverage YOLO \cite{yolov8_ultralytics} and employ our semantic lifting in clustering (center). Our path planning can leverage these semantic targets (right).}
    \label{fig:arch}
\end{figure}

\paragraph{2D Semantic Inference} 

For every frame, the detection model performs inference, outputting a set of bounding boxes $B = \{b_1, b_2, \dots, b_n\}$. Each bounding box is associated with a semantic class label from the predefined language prompts and a corresponding confidence score. 

\paragraph{Semantic Lifting and 3D Projection} 

To transition from the image plane to the world coordinate system, we perform geometric back-projection. For each 2D detection, depth values are sampled from GSFusion's fused depth map \cite{wei2024gsfusion}. Each pixel $(u, v)$ within a detection box is mapped to a 3D point $\mathbf{p}_W$ in the world frame:
$$
    \mathbf{p}_W = \mathbf{T}_{WB} \left( D(u, v) \mathbf{K}^{-1} [u, v, 1]^T \right)
$$
where $D(u, v)$ is the depth value, $\mathbf{K}$ is the camera intrinsic matrix, and $\mathbf{T}_{WB}$ is the current estimated camera pose. To mitigate boundary depth outliers, we use the median 3D position of the masked region as the detection’s centroid in the world frame. Operating online, the system observes objects from multiple viewpoints, accumulating a dense semantic point cloud over the trajectory.

\paragraph{Spatial Consolidation and Clustering} 

To mitigate projection errors, we apply \ac{dbscan} \cite{ester1996dbscan} to cluster points into distinct object instances based on Euclidean distance. Sparse points are rejected as noise, preventing false positives. Each valid cluster is assigned a centroid and unique ID, forming 3D semantic waypoints for path planning.

\subsection{Continuous-Time Trajectory Optimization}

With 3D target coordinates established, our path planner generates collision-free trajectories within the safe, tangible geometry of the \ac{tsdf}. We employ our \approach, building upon the B-Spline planner  \cite{adamkiewicz2022nerfnav,chen2025splat}. We substantially modify this baseline to operate on explicit \ac{tsdf} geometry rather than implicit radiance fields, and introduce custom boundary conditions to support robust semantic targeting.

\begin{table*}[t!]
\centering
\caption{Comparative Benchmark: Safety, Efficiency, and Trajectory Quality across diverse indoor scenes.}
\label{tab:benchmark_combined}
\resizebox{\textwidth}{!}{%
\begin{tabular}{@{}lcccccccccccccc@{}}
\hline \hline
\multirow{2}{*}{\textbf{Scene}} & \multicolumn{2}{c}{\textbf{Feasibility (\%) $\uparrow$}} & \multicolumn{2}{c}{\textbf{Success (GT) (\%) $\uparrow$}} & \multicolumn{2}{c}{\textbf{Time (s) $\downarrow$}} & \multicolumn{2}{c}{\textbf{Margin (m) $\uparrow$}} & \multicolumn{2}{c}{\textbf{Path Len (m) $\downarrow$}} & \multicolumn{2}{c}{\textbf{LE-Len (m) $\downarrow$}} & \multicolumn{2}{c}{\textbf{Max Jerk ($m/s^3$) $\downarrow$}} \\ \cline{2-15} 
 & \textit{SplatNav} & \textit{\approach} & \textit{SplatNav} & \textit{\approach} & \textit{SplatNav} & \textit{\approach} & \textit{SplatNav} & \textit{\approach} & \textit{SplatNav} & \textit{\approach} & \textit{SplatNav} & \textit{\approach} & \textit{SplatNav} & \textit{\approach} \\ \hline 
office0 & 80$\pm$40 & \textbf{100$\pm$0} & \textbf{100$\pm$0} & \textbf{100$\pm$0} & \textbf{0.15$\pm$0.03} & 0.35$\pm$0.01 & 0.11$\pm$0.07 & \textbf{0.11$\pm$0.04} & 3.74$\pm$0.73 & \textbf{2.77$\pm$0.81} & 4.30$\pm$0.72 & \textbf{3.45$\pm$0.82} & 0.65$\pm$0.17 & \textbf{0.09$\pm$0.05} \\
office2 & \textbf{100$\pm$0} & \textbf{100$\pm$0} & \textbf{100$\pm$0} & \textbf{100$\pm$0} & \textbf{0.24$\pm$0.12} & 0.40$\pm$0.14 & \textbf{0.09$\pm$0.00} & -0.02$\pm$0.09 & 4.48$\pm$2.58 & \textbf{3.40$\pm$1.48} & 5.84$\pm$2.59 & \textbf{4.07$\pm$1.52} & 0.74$\pm$0.47 & \textbf{0.11$\pm$0.03} \\
office3 & 0$\pm$0 & \textbf{100$\pm$0} & N/A & \textbf{100$\pm$0} & N/A & \textbf{0.36$\pm$0.00} & N/A & \textbf{0.16$\pm$0.04} & N/A & \textbf{4.86$\pm$0.97} & N/A & \textbf{5.56$\pm$0.98} & N/A & \textbf{0.20$\pm$0.07} \\
office4 & \textbf{100$\pm$0} & \textbf{100$\pm$0} & \textbf{100$\pm$0} & \textbf{100$\pm$0} & \textbf{0.27$\pm$0.06} & 0.36$\pm$0.00 & \textbf{0.10$\pm$0.07} & 0.01$\pm$0.07 & 6.75$\pm$1.62 & \textbf{4.85$\pm$0.92} & 7.60$\pm$1.75 & \textbf{5.53$\pm$0.92} & 1.14$\pm$0.08 & \textbf{0.17$\pm$0.05} \\
room0   & \textbf{100$\pm$0} & \textbf{100$\pm$0} & \textbf{100$\pm$0} & \textbf{100$\pm$0} & \textbf{0.36$\pm$0.08} & \textbf{0.36$\pm$0.01} & \textbf{0.13$\pm$0.06} & 0.10$\pm$0.07 & 5.98$\pm$1.39 & \textbf{4.57$\pm$0.89} & 6.79$\pm$1.35 & \textbf{5.24$\pm$0.91} & 0.76$\pm$0.07 & \textbf{0.21$\pm$0.09} \\
room1   & \textbf{100$\pm$0} & \textbf{100$\pm$0} & \textbf{100$\pm$0} & 60$\pm$49 & \textbf{0.26$\pm$0.09} & 0.35$\pm$0.00 & \textbf{0.09$\pm$0.04} & 0.02$\pm$0.09 & 3.85$\pm$1.59 & \textbf{2.86$\pm$1.10} & 4.57$\pm$1.56 & \textbf{3.52$\pm$1.12} & 0.43$\pm$0.10 & \textbf{0.11$\pm$0.02} \\
room2   & \textbf{100$\pm$0} & \textbf{100$\pm$0} & \textbf{100$\pm$0} & \textbf{100$\pm$0} & \textbf{0.24$\pm$0.05} & 0.35$\pm$0.00 & 0.10$\pm$0.09 & \textbf{0.11$\pm$0.11} & 4.41$\pm$1.14 & \textbf{3.78$\pm$1.24} & 5.27$\pm$1.30 & \textbf{4.45$\pm$1.26} & 0.58$\pm$0.10 & \textbf{0.13$\pm$0.06} \\ \hline 
\textbf{AVERAGE} & 83 & \textbf{100} & 100* & \textbf{94} & \textbf{0.25} & 0.36 & \textbf{0.10} & 0.07 & 4.87* & \textbf{3.87} & 5.73* & \textbf{4.55} & 0.72* & \textbf{0.15} \\ \hline \hline
\multicolumn{15}{l}{\footnotesize \textit{*SplatNav averages exclude office3 where the planner failed to find a valid trajectory. Success rate is calculated based on feasible paths.}}
\end{tabular}%
}
\end{table*}

\begin{table*}[t!]
\centering
\caption{Semantic Instance Targeting: Safety, Efficiency, and Trajectory Quality.}
\label{tab:semantic_combined}
\resizebox{\textwidth}{!}{%
\begin{tabular}{@{}lcccccccccccccc@{}}
\hline \hline
\multirow{2}{*}{\textbf{Scene}} & \multicolumn{2}{c}{\textbf{Feasibility (\%) $\uparrow$}} & \multicolumn{2}{c}{\textbf{Success (GT) (\%) $\uparrow$}} & \multicolumn{2}{c}{\textbf{Time (s) $\downarrow$}} & \multicolumn{2}{c}{\textbf{Margin (m) $\uparrow$}} & \multicolumn{2}{c}{\textbf{Path Len (m) $\downarrow$}} & \multicolumn{2}{c}{\textbf{LE-Len (m) $\downarrow$}} & \multicolumn{2}{c}{\textbf{Max Jerk ($m/s^3$) $\downarrow$}} \\ \cline{2-15} 
 & \textit{SplatNav} & \textit{\approach} & \textit{SplatNav} & \textit{\approach} & \textit{SplatNav} & \textit{\approach} & \textit{SplatNav} & \textit{\approach} & \textit{SplatNav} & \textit{\approach} & \textit{SplatNav} & \textit{\approach} & \textit{SplatNav} & \textit{\approach} \\ \hline 
office0 & 44$\pm$50 & \textbf{100$\pm$0} & \textbf{94$\pm$24} & 90$\pm$30 & \textbf{0.14$\pm$0.04} & 0.37$\pm$0.01 & \textbf{0.13$\pm$0.08} & 0.01$\pm$0.06 & 2.98$\pm$1.03 & \textbf{2.00$\pm$0.70} & 3.74$\pm$1.08 & \textbf{2.65$\pm$0.71} & 0.59$\pm$0.15 & \textbf{0.10$\pm$0.07} \\
office2 & 86$\pm$35 & \textbf{100$\pm$0} & \textbf{100$\pm$0} & 76$\pm$43 & \textbf{0.16$\pm$0.07} & 0.36$\pm$0.01 & \textbf{0.30$\pm$0.07} & -0.00$\pm$0.08 & 2.85$\pm$1.27 & \textbf{2.26$\pm$1.02} & 3.76$\pm$1.27 & \textbf{2.90$\pm$1.03} & 0.64$\pm$0.30 & \textbf{0.11$\pm$0.05} \\
office3 & 41$\pm$49 & \textbf{100$\pm$0} & \textbf{100$\pm$0} & 59$\pm$49 & \textbf{0.09$\pm$0.03} & 0.36$\pm$0.01 & \textbf{0.21$\pm$0.10} & 0.05$\pm$0.05 & \textbf{1.21$\pm$0.68} & 2.48$\pm$1.28 & 3.88$\pm$0.57 & \textbf{3.14$\pm$1.30} & 0.33$\pm$0.30 & \textbf{0.12$\pm$0.08} \\
office4 & \textbf{100$\pm$0} & \textbf{100$\pm$0} & \textbf{100$\pm$0} & 88$\pm$33 & \textbf{0.15$\pm$0.05} & 0.36$\pm$0.01 & \textbf{0.20$\pm$0.09} & 0.03$\pm$0.08 & 3.58$\pm$1.34 & \textbf{2.34$\pm$0.97} & 4.38$\pm$1.41 & \textbf{2.99$\pm$0.99} & 1.03$\pm$0.21 & \textbf{0.10$\pm$0.05} \\
room0   & 92$\pm$28 & \textbf{100$\pm$0} & \textbf{100$\pm$0} & 81$\pm$40 & \textbf{0.21$\pm$0.08} & 0.36$\pm$0.01 & \textbf{0.17$\pm$0.09} & 0.02$\pm$0.05 & 3.36$\pm$1.42 & \textbf{2.29$\pm$1.09} & 4.08$\pm$1.46 & \textbf{2.94$\pm$1.11} & 0.63$\pm$0.17 & \textbf{0.09$\pm$0.05} \\
room1   & \textbf{100$\pm$0} & \textbf{100$\pm$0} & \textbf{100$\pm$0} & 93$\pm$25 & \textbf{0.20$\pm$0.07} & 0.36$\pm$0.01 & \textbf{0.20$\pm$0.09} & -0.01$\pm$0.11 & 3.06$\pm$1.19 & \textbf{2.01$\pm$0.84} & 3.88$\pm$1.28 & \textbf{2.67$\pm$0.86} & 0.57$\pm$0.15 & \textbf{0.09$\pm$0.05} \\
room2   & \textbf{100$\pm$0} & \textbf{100$\pm$0} & \textbf{100$\pm$0} & 67$\pm$47 & \textbf{0.15$\pm$0.04} & 0.37$\pm$0.02 & \textbf{0.18$\pm$0.07} & -0.03$\pm$0.11 & \textbf{2.59$\pm$0.85} & 2.73$\pm$2.26 & 3.44$\pm$0.94 & \textbf{3.41$\pm$2.33} & 0.57$\pm$0.12 & \textbf{0.15$\pm$0.10} \\ \hline 
\textbf{AVERAGE} & 80 & \textbf{100} & \textbf{99} & 79 & \textbf{0.16} & 0.36 & \textbf{0.20} & 0.01 & 2.80 & \textbf{2.30} & 3.88 & \textbf{2.96} & 0.62 & \textbf{0.11} \\ \hline \hline
\end{tabular}%
}
\end{table*}

\paragraph{Trajectory Parameterization} 

The trajectory is represented as a uniform B-Spline of degree $p$ (typically $p=8$), defined by a sequence of control points $\mathbf{Q} = \{\mathbf{q}_0, \mathbf{q}_1, \dots, \mathbf{q}_n\}$. This representation guarantees $C^{p-1}$ continuity, ensuring that velocity, acceleration, and jerk are smooth. The local support of B-splines lets the optimizer adjust segments to avoid obstacles without affecting the global trajectory.

\paragraph{Robust Initialization via Discretized Search} 

To avoid local minima, we employ a hierarchical initialization. The environment is first discretized into a 3D occupancy grid with a conservative obstacle inflation layer of $1.5 \times r_{robot}$, where $r_{robot}$ is the robot's physical radius. If a requested start or goal point lies within an inflated obstacle, a nearest-neighbor search projects the point to the closest safe navigable voxel. A 3D A* search is then executed between these safe projected voxels to find a collision-free ``skeleton'' path, which is resampled to seed the initial B-Spline control points.

\paragraph{Multi-Objective Trajectory Optimization} 

We formulate path planning as a constrained optimization problem, solved iteratively using the Adam optimizer and PyTorch’s autograd engine. The objective function $J(\mathbf{Q})$ minimizes a weighted sum of four costs:
$$
    J(\mathbf{Q}) = w_{coll} L_{coll} + w_{dist} L_{dist} + w_{acc} L_{acc} + w_{jerk} L_{jerk}
$$
For collision avoidance, we leverage the implicit distance information of the \ac{tsdf} using a composite loss formulation: 
$$
    L_{coll} = \sum \left[ \max\left(0, 1 - \frac{d(\mathbf{p})}{m}\right) + \alpha \exp(-\beta d(\mathbf{p})) \right]
$$   
where $d(\mathbf{p})$ is the distance to the nearest surface sampled from the \ac{tsdf}, and $m$ is the safety margin (set to $1.5 \times r_{robot}$). The primary hinge loss component creates a potential field that aggressively pushes the path away from surfaces when they infringe upon the safety buffer.The secondary exponential term ($\alpha = 0.01, \beta = 2.0$) maintains a small, non-zero gradient outside the margin, ensuring continuous optimization.
Path efficiency is enforced by $L_{dist}$, which minimizes the squared Euclidean distance between consecutive spline points. $L_{acc}$ and $L_{jerk}$ penalize the squared magnitudes of acceleration ($\ddot{\mathbf{p}}$) and jerk ($\dddot{\mathbf{p}}$), yielding graceful motion profiles that respect the actuators' dynamical limits.

\paragraph{Convergence and Finalization} 

While the initial A* seed is bounded by the safely projected free voxels, the continuous B-Spline optimization actively targets the \textit{actual} requested start and goal coordinates. However, because semantic target coordinates inherently lie inside physical obstacle volumes (e.g., the geometric centroid of a designated object), allowing the continuous optimizer to reach the exact target would introduce inevitable collisions at the trajectory's extremities. To resolve this, the fully optimized B-Spline path is explicitly truncated at both ends by a distance equivalent to the inflation layer ($1.5 \times r_{robot}$). The endpoints of this safely cut spline are then linearly connected back to the exact start and goal coordinates. This formulation guarantees physical safety by preventing navigation inside the obstacle, while simultaneously mirroring the free-voxel projection mechanics of the baseline method (SplatNav) to ensure a fair and consistent comparison of path-length metrics.

\section{EVALUATION}\label{section:evaluation}

We compare \approach against SplatNav \cite{chen2025splat} on Replica \cite{straub2019replica} in two settings. Scenario A, ``Discretized Goal Benchmarking'', using predefined handcrafted navigation start and goal. Scenario B, ``Semantic Instance Targeting'', assesses the fully integrated pipeline using YOLOv8n-seg model \cite{yolov8_ultralytics}.

\subsection{Implementation Details}

We built upon GSFusion's \cite{wei2024gsfusion} C++ implementation. Our semantic embedding enhanced detection model is exported to the ONNX format and executed via the OpenCV DNN module utilizing a CUDA backend. 

\paragraph{Hyperparameters}

We set B-Spline degree $p=8$, exponential decay loss for collision avoidance, collision penalty weight $w_{coll} = 2000$, distance penalty weight $w_{dist} = 1.0$ (normalized by path length), acceleration penalty weight $w_{acc} = 0.5$, jerk penalty weight $w_{jerk} = 0.5$, and a simulated robot diameter of $0.2$\,m.

\paragraph{Baselines} 

For the baseline (SplatNav), which relies on Nerfstudio \cite{tancik2023nerfstudio} to generate its \ac{gs} representation, we processed the Replica dataset with their pipeline.

\paragraph{Metrics}

We measure total wall-clock time (dataset processing (raw RGB-D streams) to the final data structures), performing three repetitions per scene to account for computational variance. We quantify performance across three axes: safety (Success Rate \ac{gt}, \textit{\ac{gs} Safety Margin}), navigation efficiency (\textit{Path Length ($L$)},  \textit{Line-of-Sight Path Length ($L_{LE}$)}), and trajectory quality (\textit{Average Jerk}, \textit{Maximum Jerk}). 

\paragraph{Hardware} 

All experiments were performed on a single workstation equipped with an NVIDIA RTX 3090 GPU and an AMD Ryzen 7 7700X processor.

\subsection{Comparative Benchmark Scenarios}

\subsubsection{Reconstruction Pipeline Efficiency}

Our underlying baseline GSFusion \cite{wei2024gsfusion} demonstrates a clear efficiency advantage, outperforming Nerfstudio's mean reconstruction time in four of the seven evaluated scenes and achieving a faster overall average ($439.44$s vs. $443.30$s).  
Nerfstudio outputs solely produce \ac{gs}, forcing SplatNav to subsequently conduct expensive convex hull derivations and spatial decompositions during the active planning phase. In contrast, GSFusion simultaneously reconstructs both the \ac{gs} and the dense, obstacle-aware \ac{tsdf}. 

\subsubsection{Scenario A: Discretized Goal Navigation}

\begin{figure}[t!]
     \centering
     \begin{subfigure}[b]{0.49\columnwidth}
         \centering
         \includegraphics[width=\textwidth]{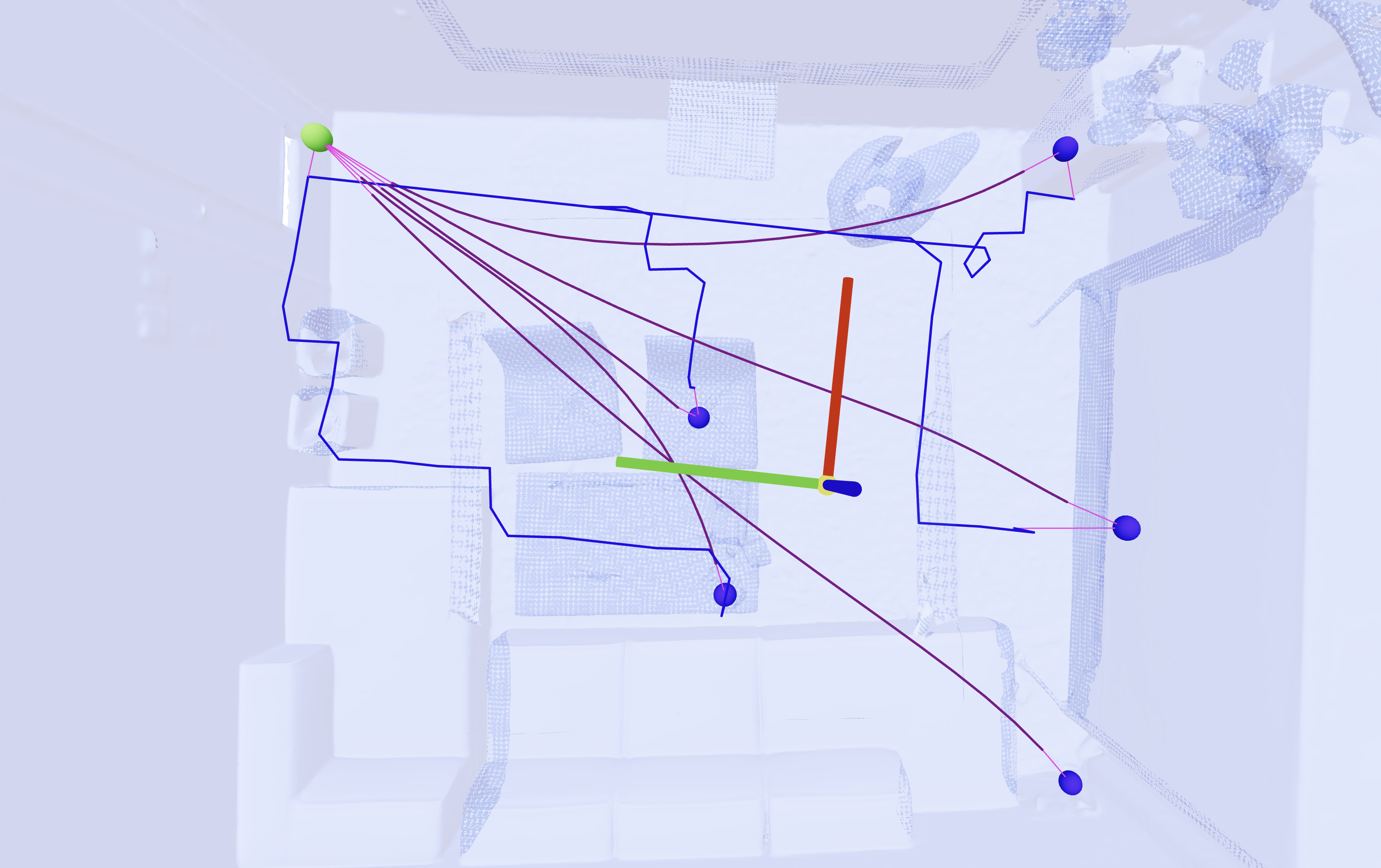}
         \caption{Top-down view}
     \end{subfigure}
     \hfill
     \begin{subfigure}[b]{0.49\columnwidth}
         \centering
         \includegraphics[width=\textwidth]{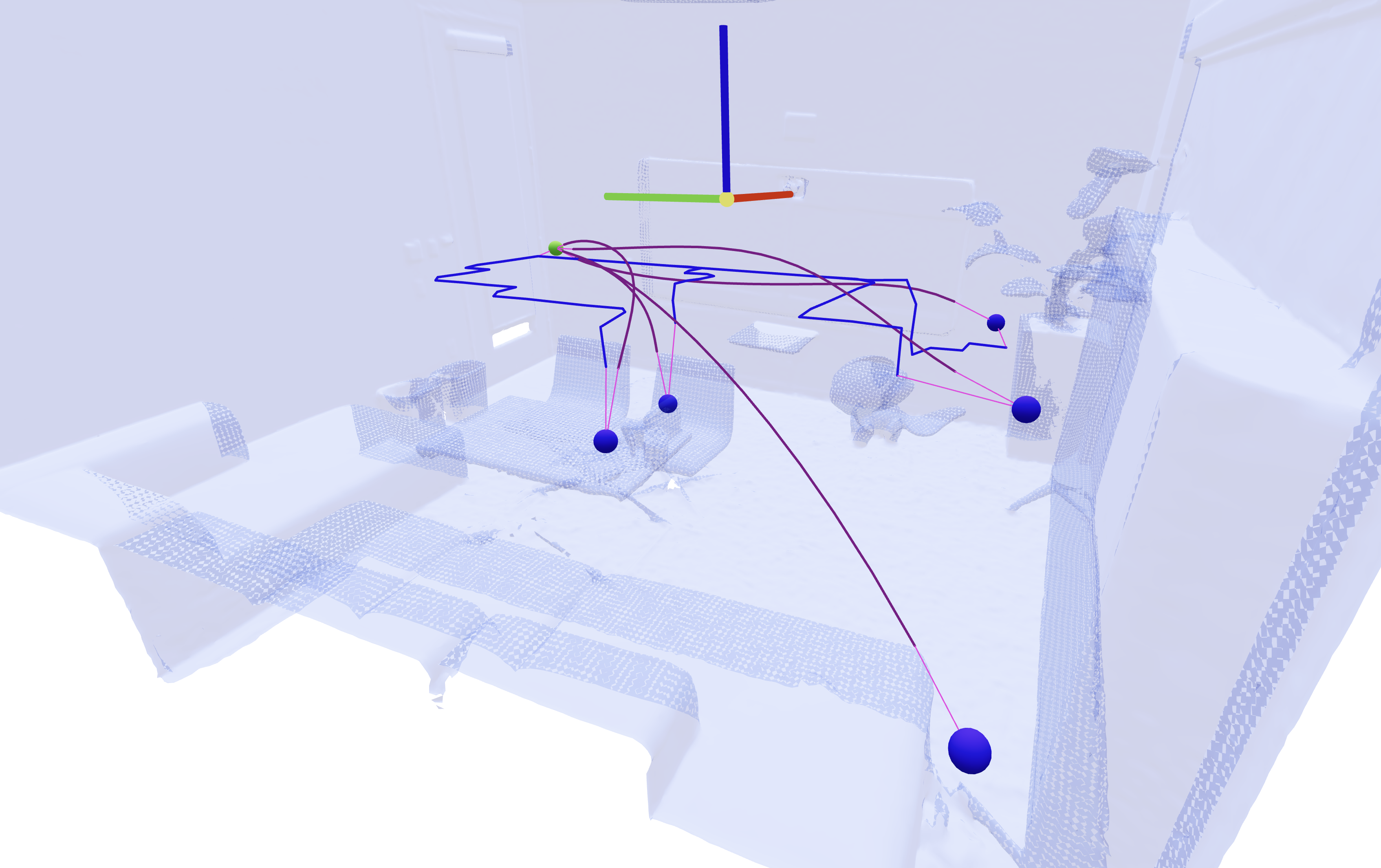}
         \caption{Side view}
     \end{subfigure}
        
     \caption{Planning results of office0 for handcrafted targets$^{1}$.}
     \label{fig:vis_comp-bench_office0}
\end{figure}

SplatNav holds a slight edge in computation time ($0.25$s vs $0.36$s) due to its convex \ac{sfc} generation (Table \ref{tab:benchmark_combined}). However, \approach compensates for this with shorter and more direct routes (Average LE-Len: $4.55$m vs $5.73$m). The smoother, shorter paths are evident on visual inspection (Fig. \ref{fig:vis_comp-bench_office0}).

\begin{figure*}[t!]
     \centering
     \begin{subfigure}[b]{0.49\columnwidth}
         \centering
         \includegraphics[width=\columnwidth]{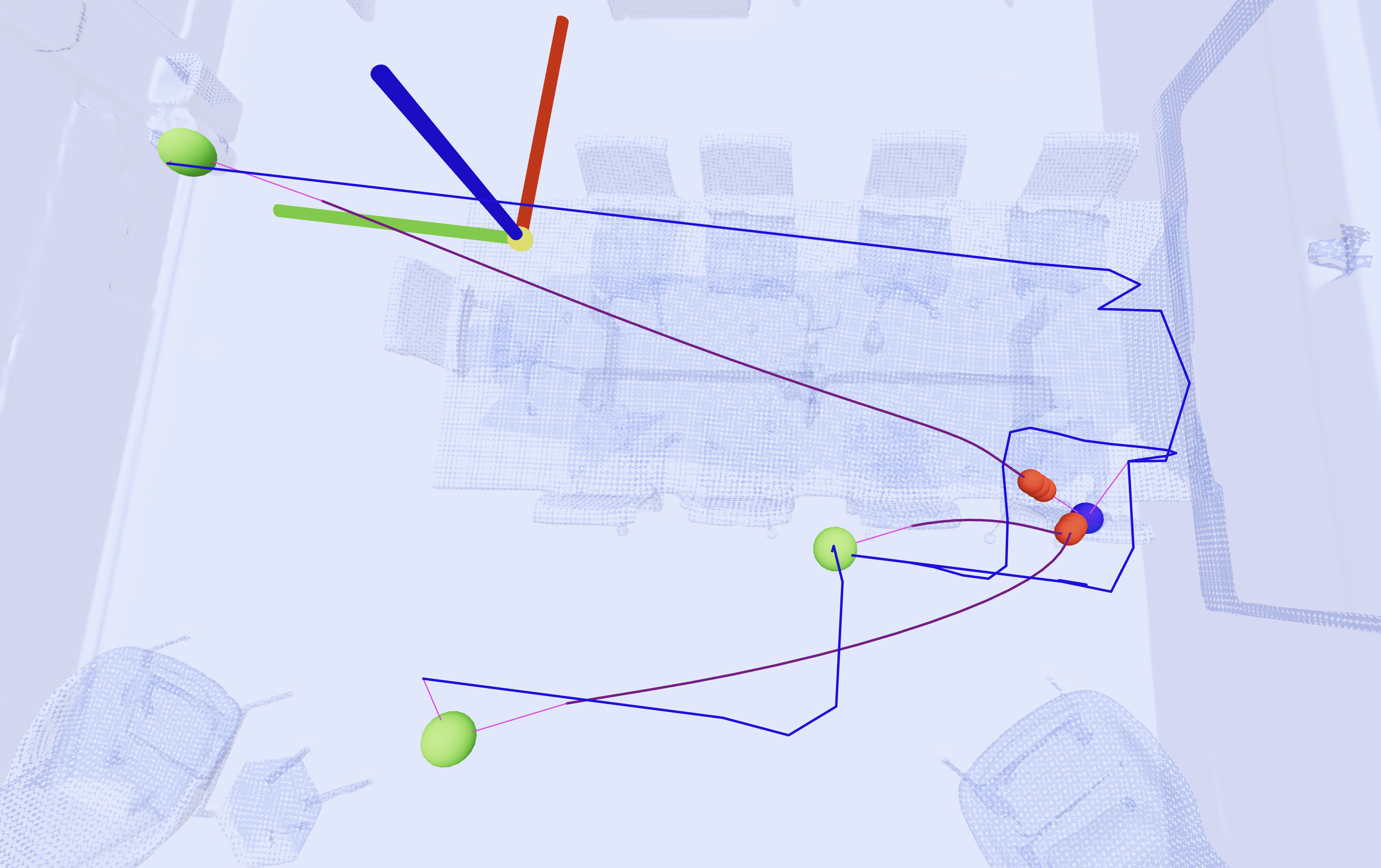}
         \caption{Top-down view}
     \end{subfigure}
     \hfill
     \begin{subfigure}[b]{0.49\columnwidth}
         \centering
         \includegraphics[width=\columnwidth]{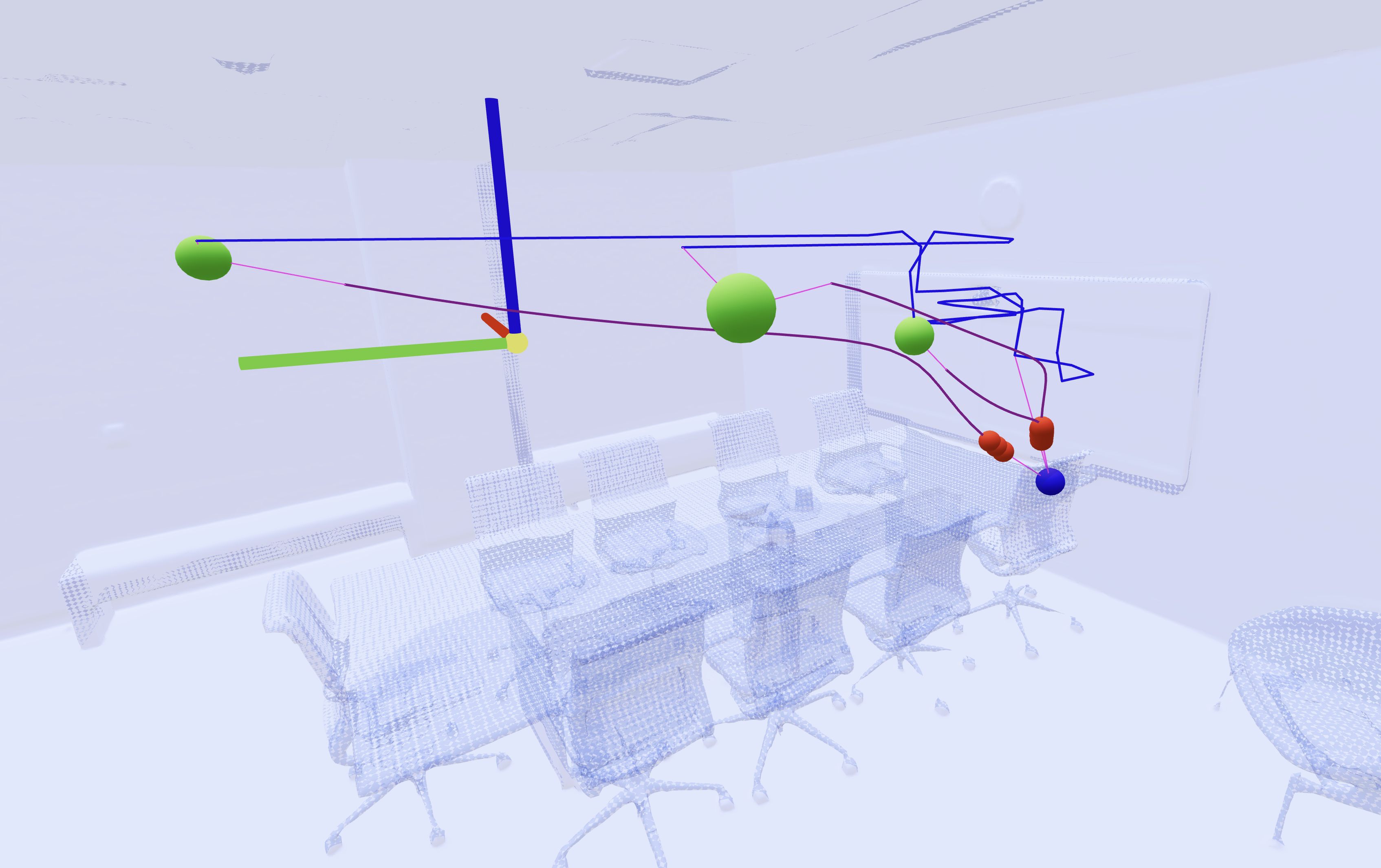}
         \caption{Side view}
     \end{subfigure}
     \hfill
    \begin{subfigure}[b]{0.49\columnwidth}
         \centering
         \includegraphics[width=\textwidth]{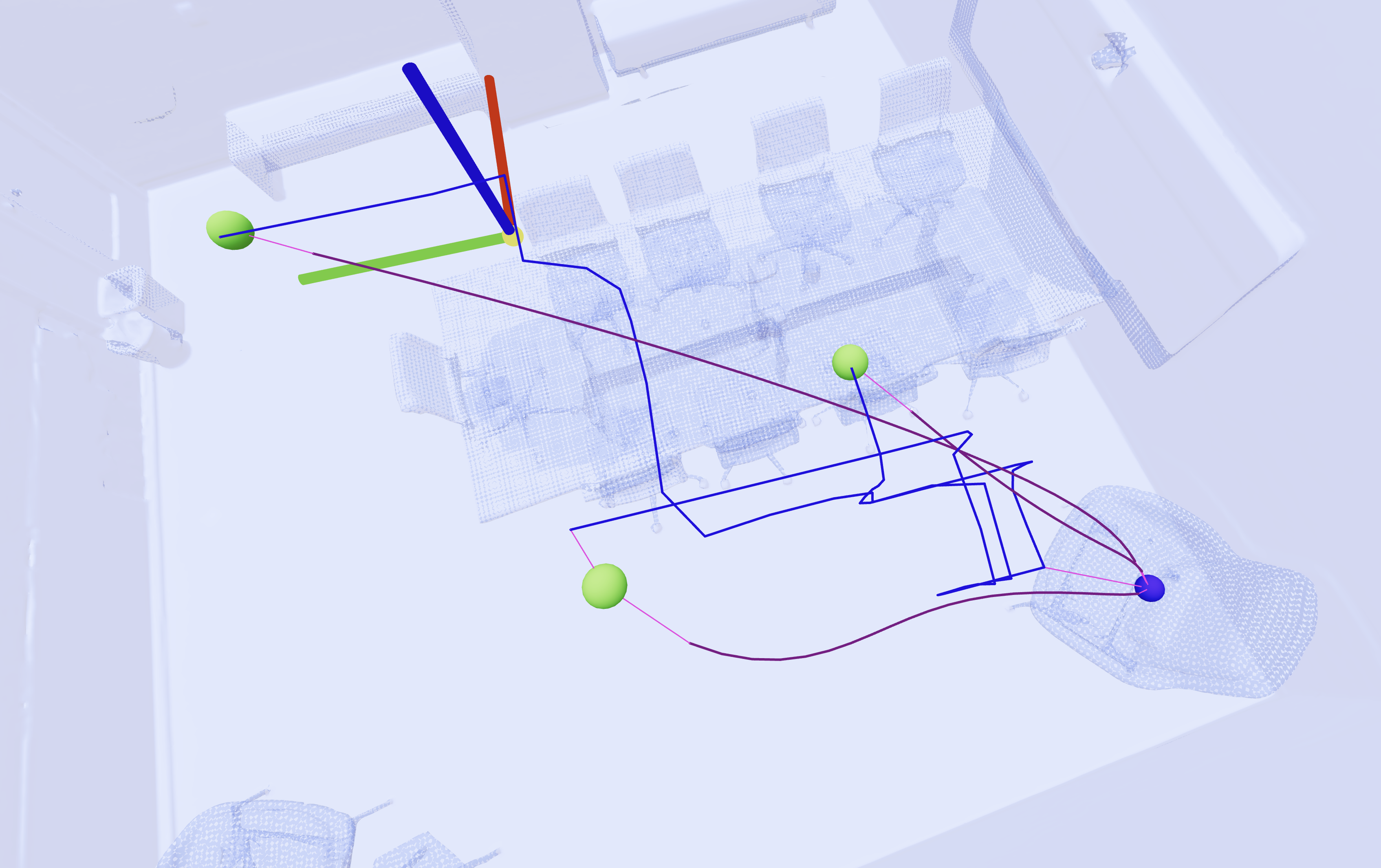}
         \caption{Top-down view }
     \end{subfigure}
     \hfill
     \begin{subfigure}[b]{0.49\columnwidth}
         \centering
         \includegraphics[width=\textwidth]{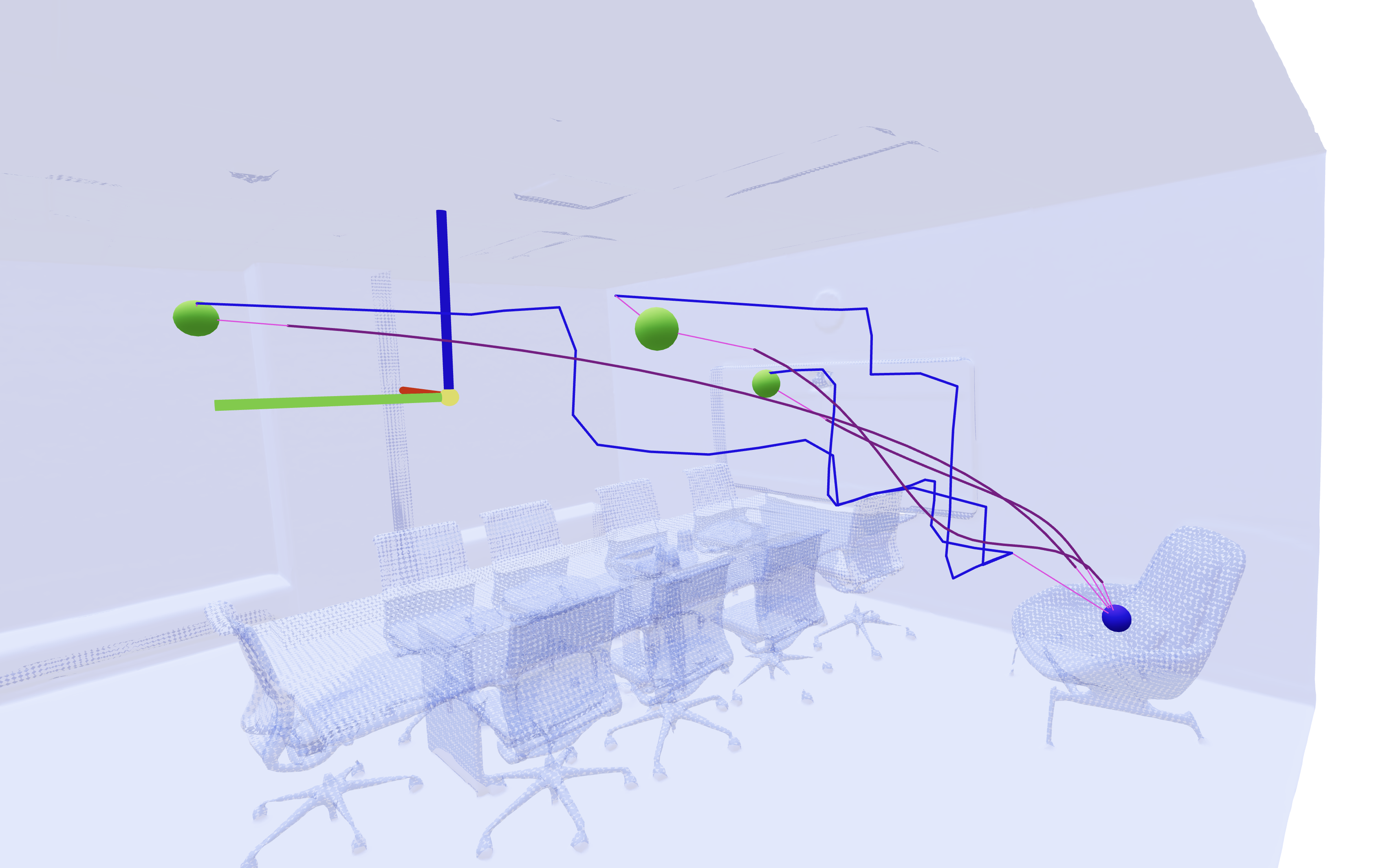}
         \caption{Side view}
     \end{subfigure}
        
     \caption{Planning results of office4 for a semantic target from the reconstruction showing collisions near the target (left, left center) and collision-free (right center, right) (SplatNav (blue), \approach (purple), start (green), target (blue), collisions (red).}
     \label{fig:office4}
\end{figure*}

\subsubsection{Scenario B: Semantic Instance Targeting}

Navigating directly to semantic objects introduces the "Semantic Trap": goal centroids often lie flush against a physical mesh (e.g., a television screen or vase). As seen in Table \ref{tab:semantic_combined}, SplatNav's \acp{sfc} struggles immensely in these scenarios, dropping to $44\%$ and $41\%$ feasibility in \textit{office0} and \textit{office3}, respectively, as it cannot mathematically generate a collision-free corridor directly into a surface boundary.
\approach, conversely, manages to achieve $100\%$ feasibility rate for semantic targeting. To reach these flush targets, our planner naturally dips into the defined safety margin at the very end of the trajectory, resulting in average margins near zero ($0.01$m) and triggering technical \ac{gt} collisions (dropping the strict success rate to $79\%$) (Fig. \ref{fig:office4}). 
Despite this strict geometric penalty at the target node, the overarching trajectory quality remains dominant (Fig. \ref{fig:office4}). \approach consistently generates shorter paths ($2.96$m vs $3.88$m LE-Len) and maintains an ultra-smooth Maximum Jerk profile ($0.11$ $m/s^3$ vs $0.62$ $m/s^3$), demonstrating its reliability and grace in fully integrated, vision-driven navigation tasks.

\section{DISCUSSION}\label{section:discussion}

While SplatNav achieved faster planning times ($\sim 0.25$s vs. $0.36$s), this speed comes at a severe cost to trajectory quality.
\approach's $C^5$ continuous B-Spline parameterization naturally regularizes the motion, suppressing peak jerk to an average of just $0.15$ $m/s^3$. 

The absolute $100\%$ feasibility rate of \approach highlights the resilience of gradient-based collision avoidance. In environments like \textit{office3}, SplatNav failed due to path initialization and subsequent optimization failures. \approach bypasses this limitation, leveraging our proposed Hinge-Loss formulation as a non-vanishing restorative force, allowing the Adam optimizer to smoothly slide control points around obstacles without becoming trapped by the topological constraints that might paralyze \ac{sfc} methods.

Scenario B shows that SplatNav struggles with semantic objects, with feasibility dropping (e.g., $41\%$ in \textit{office3}) since it cannot generate a collision-free corridor into a solid boundary. \approach succeeded in $100\%$ of these tasks. However, the data shows that to reach these targets, the optimizer deliberately forces the terminal segment of the spline into the safety margin, resulting in \ac{gt} collisions (dropping the strict success rate to $79\%$) (see Section \ref{sec:limitation}).

\section{LIMITATION and FUTURE WORK}
\label{sec:limitation}
Boundary violations originate upstream, not in the planner. Multi-view semantic lifting places target centroids deep inside objects, especially for larger objects and diverse views, forcing the optimizer to route the spline through the safety margin. This trade-off, is a design choice to focus first on efficient target oriented path planning while future work will focus on improving in-flights safety  and collision avoidance.

Currently, we only evaluate on the Replica dataset. Future work shall move to more challenging datasets and outdoor scenarios.

\section{CONCLUSION}\label{section:conclusion}

\approach introduces a hybrid navigation framework that bridges collision-safe geometric planning and flexible semantic reasoning. Built on GSFusion, it integrates semantic lifting to localize and navigate to semantic targets without dense 3D semantic mapping.
Trajectory generation is formulated as continuous B-spline optimization over \ac{tsdf} gradients, yielding smoother and more feasible motion than discretized \ac{sfc}. A hinge-loss collision term further improves the safety and efficiency trade-off by enabling smooth navigation through complex environments. The framework demonstrates strong robustness in constrained semantic trap scenarios where \ac{sfc} methods fail, providing topologically consistent and dynamically feasible trajectories.




\section*{ACKNOWLEDGMENTS}

This work was partially supported by the Technical University of Munich, Munich, Germany, through its MIRMI Seed Fund program, Huawei Technologies, Düsseldorf, Germany, and by the Robotics Institute Germany (RIG) under grant 16ME0997K.

\printbibliography

\end{document}